\documentclass[10pt, a4paper]{article}

\usepackage{lrec-coling2024} 

\usepackage{natbib}
\usepackage{multibib}
\usepackage{graphicx}
\usepackage{tabularx}
\usepackage{pdflscape}
\usepackage{multirow}
\usepackage[normalem]{ulem}
\useunder{\uline}{\ul}{}
\usepackage{lscape}
\usepackage{longtable}
\usepackage{soul}
\usepackage{float}
\usepackage{color}

\usepackage{xcolor}
\usepackage{hyperref}
 \definecolor{darkblue}{rgb}{0, 0, 0.5}
  \hypersetup{colorlinks=true, citecolor=darkblue, linkcolor=darkblue, urlcolor=darkblue}

\usepackage{xstring}

\usepackage{color}

\title{Malaysian English News Decoded: A Linguistic Resource for Named Entity and Relation Extraction}

\name{Mohan Raj Chanthran$^1$, Lay-Ki Soon$^{1*}$, Huey Fang Ong$^1$, and Bhawani Selvaretnam$^2$} 

\address{$^1$School of Information Technology, Monash University, Malaysia, $^2$Valiantlytix Sdn Bhd\\
         $^1$\{mohan.chanthran, soon.layki, ong.hueyfang\}@monash.edu, \\$^2${bhawani@valiantlytix.com}}
         
\abstract{
Standard English and Malaysian English exhibit notable differences, posing challenges for natural language processing (NLP) tasks on Malaysian English. Unfortunately, most of the existing datasets are mainly based on standard English and therefore inadequate for improving NLP tasks in Malaysian English. An experiment using state-of-the-art Named Entity Recognition (NER) solutions on Malaysian English news articles highlights that they cannot handle morphosyntactic variations in Malaysian English.  To the best of our knowledge, there is no annotated dataset available to improvise the model. To address these issues, we constructed a Malaysian English News (MEN) dataset, which contains 200 news articles that are manually annotated with entities and relations. We then fine-tuned the spaCy NER tool and validated that having a dataset tailor-made for Malaysian English could improve the performance of NER in Malaysian English significantly. This paper presents our effort in the data acquisition, annotation methodology, and thorough analysis of the annotated dataset. To validate the quality of the annotation, inter-annotator agreement was used, followed by adjudication of disagreements by a subject matter expert. Upon completion of these tasks, we managed to develop a dataset with 6,061 entities and 3,268 relation instances. Finally, we discuss on spaCy fine-tuning setup and analysis on the NER performance. This unique dataset will contribute significantly to the advancement of NLP research in Malaysian English, allowing researchers to accelerate their progress, particularly in NER and relation extraction. The dataset and annotation guideline has been published on Github. 
 \\ \newline  }

\graphicspath{ {./figures/} }
\begin{document}

\maketitleabstract
\def\thefootnote{*}\footnotetext{Corresponding Author.}\def\thefootnote{\arabic{footnote}}
\section{Introduction}
\label{sec:introduction}

\subsection{Overview}
\label{ssec:overview}
Relation Extraction (RE) is a natural language processing (NLP) task that involves identifying relations between a pair of entities mentioned in a text. This task requires identifying the entities and predicting the nature of their relationship based on the context of the sentence or document. Many previous studies on RE are based on supervised learning \citep{swampillai-stevenson-2011-extracting, chan-roth-2011-exploiting, sahu-etal-2019-inter,wang-etal-2020-global}. This means that the performance of the RE models depends very much on the quality of the annotated dataset used for the training. 

Malaysian English (ME) is a variant of English that has evolved from Standard English, incorporating local words and grammatical structures commonly used by Malaysians \citep{malaysian-english-versus-standard-english}. It is widely used in daily communications, both in informal and formal settings, such as news reports \citep{malaysian-english-versus-standard-english}. Malaysian English is considered a creole language, where it has some influence of Malay, Chinese, and Tamil together with Standard English. ME includes morphosyntactic and semantical adaptations \citep{exploring-malaysian-english-newspaper-corpus}. Listed below are some examples of morphosyntactic adaptations, the translation and meaning of these words can be referred to in the Appendix:
\begin{itemize}
    \item \textbf{Loan Words: } Words adopted from Malay and Chinese. Example: \textit{nasi lemak, amah} and \textit{ang pow}. \citep{lexical-borrowing-in-me} has also given some examples like: \textit{kenduri, imam, ustaz, bumiputera, orang asli, Datuk, Makcik, Dewan Negara, Menteri Besar} and \textit{Yang di-Pertua}. 
    \item \textbf{Compound Blends: } Words combined from two different words. Example: \textit{tidak apa attitude} and \textit{Chinese sinseh}. Some examples from \citep{lexical-borrowing-in-me} are: \textit{pondok school, Orang Asli Affairs, Tabung Haji Board} and \textit{kampung house}. 
    \item \textbf{Derived Words: } A morphosyntactic adaptions to make new words. Example: \textit{datukship (adding suffix -ship), Johorean (adding suffix -ean)} and \textit{non-halal (adding prefix non)}.
\end{itemize}

Despite being widely used in Malaysia, ME has not received much attention in the field of natural language processing (NLP) and considered low resourced language. Our goal is to help develop resources and facilitate future research in this area by introducing the Malaysian English News (MEN) corpus. 

\subsection{Motivation}
\label{ssec:motivation}
As discussed in Section \ref{ssec:overview}, the differences between Standard English and Malaysian English make the data collection process challenging. Existing datasets, which are primarily based on Standard English (American or British English), are not appropriate for this study. 

An effective relation extraction model relies on accurate named entity recognition (NER). As noted in Section \ref{ssec:overview}, the entities in Malaysian English news articles exhibit morphosyntactic variations, which necessitates the expansion of existing NER solutions for accurate entity extraction. To assess the precision of existing NER solutions for Malaysian English news articles, we carried out an experiment using several selected tools, including spaCy \citep{spacy3}, Flair \citep{akbik2019flair}, Stanza \citep{stanford-stanza}, and Malaya \citep{malaya}. To ensure a fair comparison, these NER models are selected because they meet two criteria: i) It was trained using the OntoNotes 5.0 dataset, which is a widely used and comprehensive dataset for NER tasks. ii) It achieved the highest F1-Score among all the available NER models provided by the tool we used.  We selected 30 sentences from Malaysian English news articles and observed how well the NER models detect entities with morphosyntactic adaptations. 

We have conducted an experiment with 30 Malaysian English sentence, to predict entities. The 30 sentences were manually annotated and compared with the entities predicted by the NER models. The chosen NER models achieved a micro F1-Score of less than \textbf{0.6} and the best performing NER tool is spaCy with F1-Score of 0.58. Meanwhile other NER tool like Flair, Stanza, and Malaya achieved F1-Score of 0.55, 0.43 and 0.55 respectively. In particular, the four NER models exhibited low effectiveness in predicting entities from the \textit{LOC}, \textit{GPE}, and \textit{EVENT} categories. This observation strongly suggests that existing models cannot accurately predict entities in Malaysian English. 

In addition to the experimental results, our survey of existing datasets revealed that no Malaysian English dataset has ever been developed. The majority of existing RE datasets like DocRED \citep{yao-etal-2019-docred}, ACE-2005 \citep{Walker2005-ym}, TACRED \citep{zhang2017tacred}, FewRel \citep{han-etal-2018-fewrel}, and CodRED \citep{yao-etal-2021-codred} are based on the Standard English.  Taking into account these two motivations, we have created our annotated dataset explicitly tailored to the Malaysian English context. An annotated dataset in Malaysian English is crucial to handle the semantic and morphosyntactic adaptations present in Malaysian English news articles. 

\subsection{Contribution}
\label{ssec:contribution}
The main contribution of this work is a Malaysian English News  (\textbf{MEN}) dataset with annotated entities and relations. 200 Malaysian English News articles have been manually annotated by four well-trained human annotators. In total, we collected 6,061 annotated entities and 3,268 relation instances from the annotations. Based on the observation, around 60\% of the entity mention from \textit{PERSON, ORGANIZATION, ROLE, TITLE} and \textit{FACILITY} are very much localised to Malaysian contexts and share the adaptation from Bahasa Malaysia. To our knowledge, this dataset is one of its kind, focusing specifically on Malaysian English news articles. This dataset can also be used for other NLP tasks like Named Entity Recognition (NER), Relation Extraction (RE), Event Extraction (EE), Coreference Resolution and Semantic Role Labelling. 
Using with dataset, we have also fine-tuned the spaCy NER model. The importance of fine-tuning spaCy model is to find if there is any improvement in entity extraction from Malaysian English context. We evaluate the performance and how fine-tune NER mode has overcomes gaps. The dataset together with annotation guideline has been published in: \url{https://github.com/mohanraj-nlp/MEN-Dataset}

The paper is structured as follows. Section 2 provides overview of existing entity recognition and relation extraction datasets which are relevant to our work. Section 3 describes the news articles we collected for human annotation. Section 4 discusses the fine-tuning of spaCy model with MEN-Dataset and evaluate performance of NER model. Finally, Section 5 concludes our work presented in this paper and points to potential future work.

\section{Related Work}
\label{sec: related-work}

\subsection{Existing Low Resource Language NER Dataset}
\label{ssec:existing-dataset-ner}
Most of the prominent high-quality named entity datasets like The Message Understanding Conference 6 (MUC-6) \citep{grishman-sundheim-1996-message}, CoNLL-03 \citep{tjong-kim-sang-de-meulder-2003-introduction} and OntoNotes 5.0 \citep{ontonotes-new} are mainly focused on Standard English. It is important to understand the annotation methodology of low resource language dataset in order to improve our annotation work. In conjunction with that we have studied on several low resource NER dataset. 

Wojood  is a nested entity dataset developed for Arabic language \citep{jarrar2022wojood}. The entity labels to annotate Wojood are adapted based on dataset OntoNotes 5.0. To ensure the quality of annotation, the annotator has proposed to calculate Inter-Annotator Agreement (IAA) using F1-Score. The higher the F1-Score, the higher the agreement between annotators. It is reported that Wojood achieved outstanding micro F1-Score of \textbf{0.976}. \citep{buaphet-etal-2022-thai} published a large-scale Nested Named Entity Recognition (NER) dataset for one of the Asian low-resource languages, Thai. The dataset covers 10 coarse grained entity labels like PERSON, ORGANIZATION, NUMBER, WORK\_OF\_ART, NORP, MISCELLANEOUS, LOCATION, DATE, FACILITY and EVENT. Another highlight of the dataset is that it includes nested entities with maximum depth of 8 layers. The approach of labelling nested entity has been applied when annotating MEN-Dataset. 

The datasets discussed above are representative of monolithic languages, as they do not exhibit significant influences from other languages. However, in our current context, we are dealing with creole languages, where there is a notable influence of other languages with Standard English. MasakhaNER is one of the large scale creole languages NER dataset that is built on 10 under-presented African language \citep{adelani2021masakhaner}. Development of MasakhaNER has help in the development and evaluation of NER models for the 10 languages. The dataset has annotated with four entity types PERSON, ORGANIZATION, LOCATION and DATE. The annotation methodology discussed in the paper has helped us to build on our annotation guideline, with some changes based on our objective. Prior to the development of MasakhaNER, there has been development of another creole language based dataset called NaijaNER \citep{oyewusi2021naijaner}. NaijaNER is developed based on 5 Nigerian Languages (Nigerian English, Nigerian Pidgin English, Igbo, Yoruba and Hausa). NaijaNER has adapted entity labels from OntoNotes 5.0, and it consists of 18 entity labels. 

Our initial study on low resource and creole language NER datasets has given us the understanding on the data collection process, annotation methodology and gaps that has been solved by these datasets. 

\subsection{Existing Datasets for Relation Extraction}
\label{ssec:existing-dataset-re}
There are numerous Relation Extraction datasets available, where ACE-2005 stands out as one of the widely-used benchmark dataset. ACE-2005 \citep{Walker2005-ym} provides English, Arabic, and Chinese annotations and 18 relation labels. High quality and extensive annotation make ACE-2005 popular among researchers. Apart from relation, ACE-2005 also includes annotated entity and events. It may provide little detail for annotating relations spanning multiple sentences or document-level relations. 

DocRED \citep{yao-etal-2019-docred} is a popular dataset for inter-sentential relation extraction models. DocRED captures relations between entities over several sentences in a document, unlike other relation extraction datasets. In document-level relation extraction research, this dataset helps researchers capture relations across phrases.

Thorough studies of ACE-2005 and DocRED do not only provide us solid understanding of the basic approach used for annotating named entities and relations, but also relevant relation labels that can be incorporated into our dataset.

\section{Malaysian English News (MEN) Dataset with Annotated Entities and Relation}
\label{sec: mendataset}
\subsection{MEN-Dataset Acquisition}
\label{ssec: overview}
A total of 14,320 news articles were scrapped from prominent Malaysian English news portals, including New Straits Times (NST) \footnote{\url{https://www.nst.com.my/}},  Malay Mail (MM)\footnote{\url{https://www.malaymail.com/}} and Bernama English \footnote{\url{https://www.bernama.com/en/}}. This 14,320 news articles are compilled as MEN-Corpus. 
\begin{table}[H]
\centering
\begin{tabular}{|l|c|c|c|c|}
\hline
         & MM & NST  & Bernama & Total \\ \hline
Nation   & 1,938      & 3,185 & 2,890    & 8,013  \\ \hline
Business & 90        & 2,218 & 1,757    & 4,065  \\ \hline
Sports   & 61        & 1,474 & 707     & 2,242  \\ \hline
Total    & 2,089      & 6,877 & 5,354    & 14,320 \\ \hline
\end{tabular}
\caption{Number of news articles collected across news portals and categories.}
\label{tab:news-article-collected}
\end{table}
In Table \ref{tab:news-article-collected} shows the statistics of MEN-Corpus, where the articles are scrapped from news categories like Nation, Business and Sports. We have selected 200 news articles from MEN-Corpus to develop MEN-Dataset. The dataset consists of 120 news from category Nation, 60 news articles from Business and 20 from Sports. Diverse categories in the corpus enable the construction of more robust NER and RE models

\subsection{Annotation Process}
\label{ssec: annotation-process}

\subsubsection{Annotation Setup}
\label{sssec: annotation-setup}

Annotators who are proficient in both Malaysian English and Bahasa Malaysia are required for the annotation task. The annotators were selected after they went through training and assessments. Four annotators who performed well in the assessments were selected. The annotators are divided into two groups and each group is assigned to annotate 100 Malaysian English news articles within 8 weeks milestone.  

\begin{figure*}[h]
    \centering
    \frame{\includegraphics[height=\textheight,width=\textwidth,keepaspectratio]{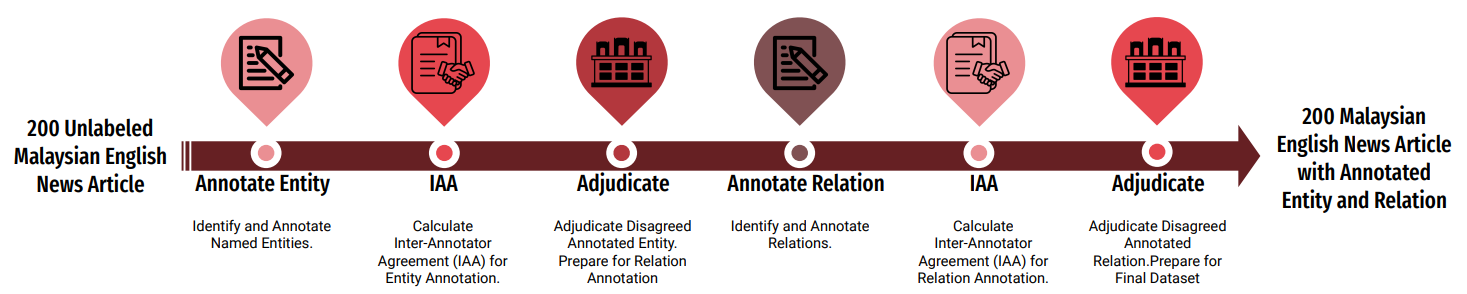}}
    \caption{Phases in the annotation process to annotate news article for each milestone. This phase has helped to ensure accuracy and consistency.}
    \label{fig:annotation-process-fig}
\end{figure*}

Figure \ref{fig:annotation-process-fig} shows the annotation process we have followed for each milestone, where the first four focus on entity annotation, and the last four on relation annotation. To maintain consistency and ensure quality, we conducted an analysis of Inter-Annotator Agreement (IAA) at each milestone. Any disagreements that arise was resolved by an adjudicator. This approach aims to prevent annotation errors and produce high-quality datasets.

The annotation guideline \footnote{\href{https://anonymous.4open.science/r/MEN-Dataset-C6F7/guideline/}{Annotation Guideline}} defines entity labels and gives examples of entity mentions. For relations, we provided the definition of relation labels and possible entity labels that can be associated with the relations. The guideline was prepared in an iterative manner, where it was first produced before the annotation started, then incrementally refined based on the annotation progress, with concise instructions.

\subsubsection{Entity Annotation}
\label{sssec: entity-annotation}

\begin{figure}[]
    \centering
    \includegraphics[width=0.5\textwidth,height=\textheight,keepaspectratio]{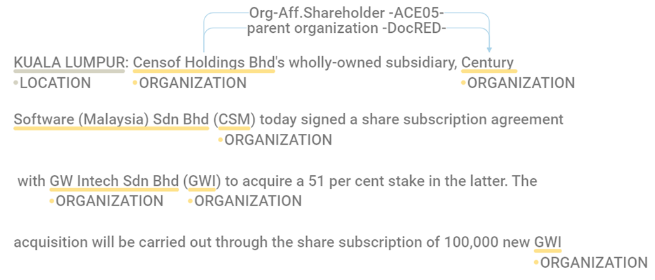}
    \caption{Snippet of news article that has been annotated with entities and relations. The annotated entities are underlined, and their corresponding entity labels are included below the line. Additionally, we linked the relations between two annotated entities. Each relation has a suffix of the dataset name, which indicates the dataset from which the relation label has been adapted.}
    \label{fig:annotation-snippet}
\end{figure}

The entity labels used for annotation are derived from OntoNotes 5.0 \citep{ontonotes-new}. OntoNotes 5.0 has 18 entity labels, which include 11 named and 7 value entities. As value entities do not require morphological or syntactic adaptation, we only included named entities in our annotation rules. The entity labels adapted and used for entity annotation are PERSON, LOCATION, ORGANIZATION, NORP, WORK\_OF\_ART, LAW, LANGUAGE, FACILITY, and PRODUCT. In order to support Malaysian English, we also included two additional entities: 
\begin{itemize}
    \item \textit{TITLE}: \textit{TITLE} refers to academic, religious and Malay royalty titles. This is added to capture the morphological adaptation, specifically for \textit{PERSON} entities that appear with their title, as part of their name in Malaysian English news. Example of entity mentions are \textit{Datuk, Datin} and \textit{Tan Sri}.
    \item \textit{ROLE}: \textit{ROLE} refers to the position that a \textit{PERSON} is holding. The \textit{ROLE} also enables us to capture the morphological adaptation. It usually comes together with the name of the \textit{PERSON}. Example of entity mentions are \textit{Mentri Besar, co-founder} and \textit{police chief ACP}.
\end{itemize}

In annotation guideline lists entity labels and descriptions. Figure \ref{fig:annotation-snippet} displays news articles annotated with \textit{ORGANIZATION} and \textit{LOCATION} labels. Once the annotators completed entity annotation, we calculated the IAA and later will be adjudicated if there is any disagreements.

\subsubsection{Relation Annotation}
\label{sssec: relation-annotation}
The relation labels for annotations are adapted from DocRED \citep{yao-etal-2019-docred} and ACE-2005 \citep{Walker2005-ym}, with some relation labels related to DATE or TIME omitted. To facilitate the annotation, we provided the annotators with a list of possible Argument Type. Argument Type aids annotators with possible entity types for each relation labels. 

In annotation guideline we have listed relation labels, descriptions, and the corresponding argument types. In Fig. \ref{fig:annotation-snippet}, a relation named \textit{manufacturer - DocRED} is used to link \textit{Honda Malaysia} and \textit{City Hatchback}. Referring to annotation guideline for relation, we can verify that the Argument Type for \textit{manufacturer - DocRED} is \textit{ORGANIZATION} and \textit{PRODUCT}. 

\subsubsection{Inter-Annotator Agreement}
\label{sssec: iaa-analysis}
In order to ensure quality annotations, Inter-Annotator Agreement (IAA), a crucial metric for evaluating annotation made by human annotators was used. The agreement was calculated by comparing the number of identical labels assigned by two or more annotators working on the same task \citep{iaa-handbook-of-linguistics-annotation}. We calculate IAA separately for entity and relation annotations as it will provide a comprehensive understanding of the accuracy and reliability of the annotations.  Cohen's Kappa was not used because it is suitable for tasks involving negative cases \citep{f-measure-and-reliability-IAA,clinical-trials-f1-score}.  In our case, there is no negative cases in entity annotations. 

In each group, one annotator is designated as the Gold Standard, and their annotations are compared to those done by the other annotator in the same group to calculate the F1-Score. Higher F1-Score indicates a higher level of agreement between the two annotators in that group \citep{f-measure-and-reliability-IAA}.

\textbf{IAA Analysis for Entity Annotation} 
Two aspects were considered when calculating F1-Score between the two annotators in each group: (i) the exact span of the entity mention and (ii) the entity label assigned to the entity mention. For entity annotations, we achieved macro F1-Score of 0.818. Several datasets, such as \citep{jarrar2022wojood, brandsen-etal-2020-creating, jiang-etal-2022-annotating}, evaluated IAA using F1-Score. \citep{jarrar2022wojood} which adapted entity labels from OntoNotes 5.0 and achieved an outstanding micro F1-Score of \textbf{0.976}. Our IAA is considered lower if compared with \citep{jarrar2022wojood}. To assure dataset quality, we adjudicated disagreed entity annotations. A subject matter expect has been appointed as adjudicator, to evaluate and make final decision on the disagreed annotations. 
\begin{figure}[!ht]
    \centering
    \includegraphics[height=\textheight,width=0.5\textwidth,keepaspectratio]{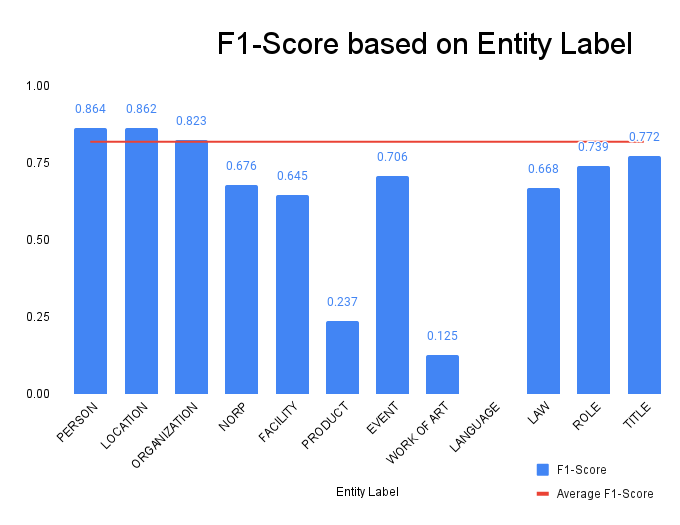}
    \caption{F1-Score calculated to measure the agreement based on entity labels.}
    \label{fig:f1-score-iaa-entity-label}
\end{figure}

Here are some observation of F1-Score from Figure \ref{fig:f1-score-iaa-entity-label}:
\begin{enumerate}
    \item For entity label \textit{LANGUAGE}, the F1-Score shows NA. This happens because there is no entity mention that belongs to that label has been annotated. More details on the total number of entities mention can be found in Figure \ref{fig:total-and-unique-entity-mention}. 
    \item \textit{PERSON} achieved the highest F1 score, which is 0.864, and the second highest is  \textit{LOCATION} with F1-Score of 0.862. 
    \item The entity label \textit{WORK\_OF\_ART} achieved the lowest F1-Score, which is 0.125. 
\end{enumerate}

If we compare the F1-Score calculated across the entity labels (Fig. \ref{fig:f1-score-iaa-entity-label}) and the total number of annotated entity mentions (Fig. \ref{fig:total-and-unique-entity-mention}), we can observe that entity labels with F1-Score below the average F1-Score have a total number of entity mentions less than 500 each. With limited examples to refer to, the annotators may have varying interpretations or perspectives on how to label entities correctly.

For instance, one of the entity mentions from entity label \textit{WORK\_OF\_ART}, \textit{Puteri Gunung Ledang} is a Fictional Character and according to the annotation guideline it should have been annotated as \textit{PERSON}. However, the annotators  annotated it as \textit{WORK\_OF\_ART} because they mistaken it for the name of a film. Another example, \textit{Most Established State in Healthcare Travel} is actually a recognition, thus the mention is not suitable to be annotated as \textit{WORK\_OF\_ART}.  This issue can be resolved during adjudication. Based on the analysis, we found that the annotators did not face any difficulty handling annotation for an entity like \textit{PERSON}, \textit{LOCATION}, and \textit{ORGANIZATION}. This is because the entity mentions that belong to these classes are clear and easily identifiable.   

\textbf{IAA Analysis for Relation Annotation} 
For IAA of Relation Annotation, we defined a criterion where both annotators have to agree on an exact match of relation instances. The macro F1-Score for overall relation annotation is \textbf{0.51}, which indicates a moderate level of agreement between the annotators in the relation annotation task. 
\begin{table}[!ht]
\begin{tabular}{|l|cl|cl|}
\hline
No  &  \multicolumn{2}{c|}{\begin{tabular}[c]{@{}c@{}}Relation Label\\ Adapted \end{tabular}}& \multicolumn{2}{c|}{\begin{tabular}[c]{@{}c@{}}IAA\\ (F1-Score)\end{tabular}} \\ \hline
1 & \multicolumn{2}{c|}{DocRED}   & \multicolumn{2}{c|}{0.567}                                                                          \\ \hline
2 & \multicolumn{2}{c|}{ACE-2005} & \multicolumn{2}{c|}{0.312}                                                                          \\ \hline
3 & \multicolumn{2}{c|}{Overall}  & \multicolumn{2}{c|}{0.512}                                                                          \\ \hline
\end{tabular}
\caption{F1-Score calculated to measure the agreement based on Relation Labels.}
\label{tab:f1-score-iaa-relation-label}
\end{table}

In Table \ref{tab:f1-score-iaa-relation-label}, we examined the IAA with relation labels adapted from DocRED and ACE-2005 respectively. Notably, the IAA for ACE-2005 is lower compared to DocRED, despite ACE-2005 having fewer relation labels (17) than DocRED (84). In this research, we have also introduced an extra relation label, \textit{\textbf{NO\_RELATION}}, which can be used when no appropriate relation label is available for a given instance. During our post-annotation feedback session, it became apparent that the annotators encountered difficulties when annotating with relation labels from ACE-2005, in contrast to their experience with DocRED. ACE-2005 has limited relation labels available for specific scenarios and argument types. This narrower scope can make it more challenging for annotators to annotate appropriate relation labels. However, for each disagreement over the annotation of the relation, the adjudicator will make the final decision.

\subsection{Dataset Statistics}
\label{ssec: dataset-stats}
In this section, we describe the dataset composition, including the distribution of annotated entities based on selected entity labels and relations. We also discuss about the entity pairs and relations occurring together, providing valuable insights into the inter-connectedness and contextual dependencies among entities and relations in MEN articles. Listed below are some statistics on the annotated MEN-Dataset:
\begin{enumerate}
    \item Total Entities of 6,061 and 2,874 unique entities. The entities are annotated based on 12 entity labels. 
    \item Total Relation Instance of 4,095. We have 2,237 and 1,031 relation instance based on labels adapted from DocRED and ACE-05 respectively. Around 827 relation instance has annotated with label NO\_RELATION.
\end{enumerate}

\paragraph{Statistics for Entity Annotation}
\begin{figure}[!ht]
    \centering
    \includegraphics[height=\textheight,width=0.5\textwidth,keepaspectratio]{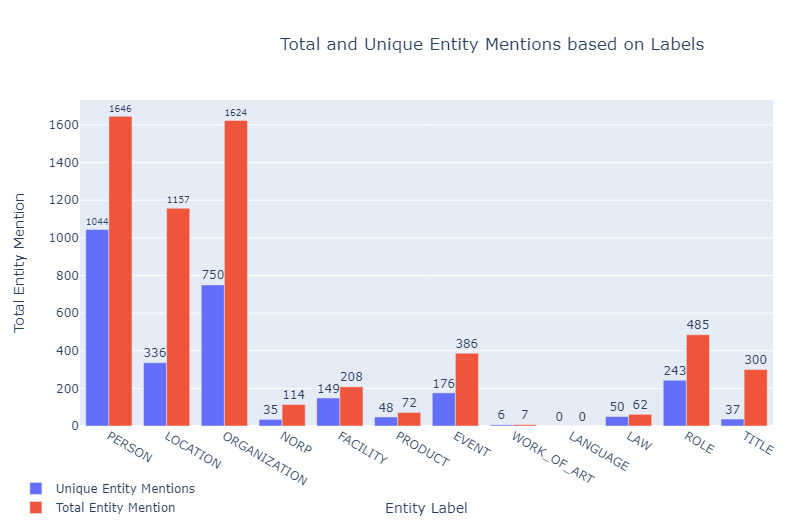}
    \caption{Total and Unique Entity Mention Annotated based on the Labels.}
    \label{fig:total-and-unique-entity-mention}
\end{figure}
Figure \ref{fig:total-and-unique-entity-mention} shows the total and unique entity mentions that have been annotated. Most of the entity mentions annotated in the news articles are \textit{PERSON, LOCATION} and \textit{ORGANIZATION}. \textit{WORK\_OF\_ART } is considered the entity label with least mentions, and there is no entity mentions belong to \textit{LANGUAGE}. 

From the 6,061 entity mentions that have been annotated, 67\% of them are in Nation category, 22\% in Business and 11\% in Sports. Based on our further analysis, around 60\% of the entity mentions from \textit{PERSON, ORGANIZATION, ROLE, TITLE} and \textit{FACILITY} are based on the context of Malaysian English and share the adaptation from Malay language. Around 42\% of 1646 entity mentions for \textit{PERSON} is associated together with either an entity mention from \textit{ROLE} or \textit{TITLE}. We would like to highlight some examples of annotated entities that exhibits morphosyntactic adaptation (we will provide more samples in Appendix): 
\begin{enumerate}
    \item PERSON, TITLE:
    \begin{enumerate}
        \item Tan Sri Dr Noor Hisham Abdullah. “Tan Sri” is a loanword, it is a common honorific title given for PERSON.
        \item Raja Permaisuri Agong:  “Raja Permaisuri Agong” is a loanword, it is a Royal TITLE given for people from royal families. 
    \end{enumerate}
    \item ORGANIZATION:
    \begin{enumerate}
        \item Bank Negara Malaysia. Bank Negara Malaysia is considered a compound blend as it combines words from both English and Bahasa Malaysia. “Bank”, representing the institution's. “Negara” is a Malay word meaning “nation”, “Malaysia” indicates geographical location. 
    \end{enumerate}
    \item NORP:
    \begin{enumerate}
        \item Sarawakians. Sarawakians is a derived word that indicates the people from state Sarawak. 
    \end{enumerate}
\end{enumerate}

\paragraph{Statistics for Relation Annotation}
Referring to dataset statistics listed in starting of Section \ref{ssec: dataset-stats}, 826 instances were labeled as \textit{NO\_RELATION}. Out of the 84 relation labels adapted from DocRED, only 51 unique relation labels were utilized for this annotation task. In contrast, for ACE-2005, 16 out of the total 17 relation labels were employed. Based on the distribution of annotated relation instances, 69\% of the instances are based on the relation labels adapted from DocRED. We have designed the guidelines to ensure that the dataset covers relations from both inter-sentential and intra-sentential contexts. Out of the 3,268 relation instances, 54\% are for intra-sentential relations, while the remaining 46\% are for inter-sentential relations.

\section{Spacy}
\label{sec: spacy}

\subsection{Background}
\label{ssec: spacy-intro}

\begin{figure}[!ht]
    \centering
    \includegraphics[width=0.5\textwidth,height=0.25\textheight,keepaspectratio]{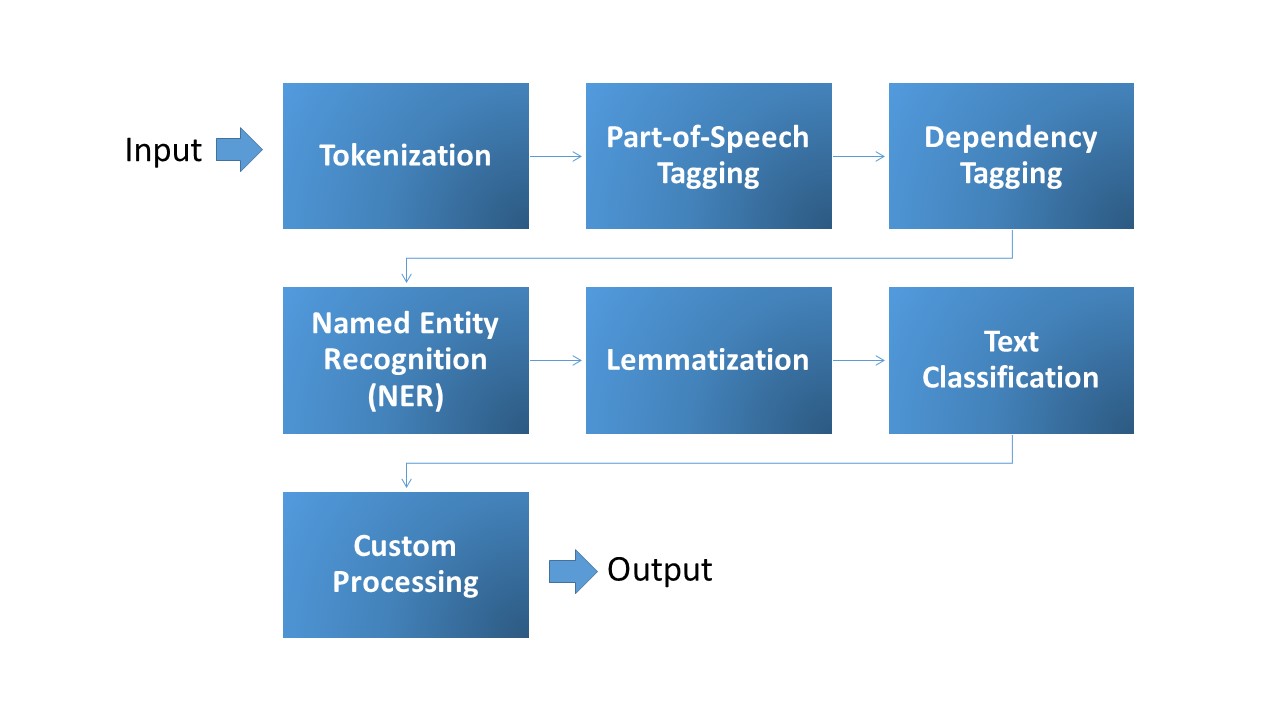}
    \caption{Proccessing steps in spaCy pipeline, it starts by giving the input. The outcome will include linguistic annotations after all the processing steps.}
    \label{fig:spacy-pipeline}
\end{figure}

\begin{table*}[!ht]
\centering
\resizebox{\textwidth}{!}{%
\begin{tabular}{|c|cc|ccccccc|}
\hline
Entity Label                                                     & \multicolumn{1}{c|}{\begin{tabular}[c]{@{}c@{}}Total Annotated \\ Entity in \\ MEN-Dataset\end{tabular}} & \begin{tabular}[c]{@{}c@{}}Total Annotated \\ Entity in \\ Validation Set\end{tabular} & \multicolumn{1}{c|}{spacy-blank} & \multicolumn{1}{c|}{\begin{tabular}[c]{@{}c@{}}en\_core\_\\ web\_sm\end{tabular}} & \multicolumn{1}{c|}{\begin{tabular}[c]{@{}c@{}}en\_core\_\\ web\_sm-FT\end{tabular}} & \multicolumn{1}{c|}{\begin{tabular}[c]{@{}c@{}}en\_core\_\\ web\_lg\end{tabular}} & \multicolumn{1}{c|}{\begin{tabular}[c]{@{}c@{}}en\_core\_\\ web\_lg-FT\end{tabular}} & \multicolumn{1}{c|}{\begin{tabular}[c]{@{}c@{}}en\_core\_\\ web\_trf\end{tabular}} & \begin{tabular}[c]{@{}c@{}}en\_core\_\\ web\_trf-FT\end{tabular} \\ \hline
PERSON                                                           & \multicolumn{1}{c|}{1646}                                                                                & 108                                                                                    & \multicolumn{1}{c|}{0.982}       & \multicolumn{1}{c|}{0.39}                                                         & \multicolumn{1}{c|}{0.852}                                                           & \multicolumn{1}{c|}{0.555}                                                        & \multicolumn{1}{c|}{0.806}                                                           & \multicolumn{1}{c|}{0.811}                                                         & 0.923                                                            \\ \hline
LOCATION                                                         & \multicolumn{1}{c|}{1157}                                                                                & 150                                                                                    & \multicolumn{1}{c|}{0.942}       & \multicolumn{1}{c|}{0.04}                                                         & \multicolumn{1}{c|}{0.897}                                                           & \multicolumn{1}{c|}{0.041}                                                        & \multicolumn{1}{c|}{0.906}                                                           & \multicolumn{1}{c|}{0.043}                                                         & 0.916                                                            \\ \hline
ORGANIZATION                                                     & \multicolumn{1}{c|}{1624}                                                                                & 262                                                                                    & \multicolumn{1}{c|}{0.956}       & \multicolumn{1}{c|}{0.474}                                                        & \multicolumn{1}{c|}{0.884}                                                           & \multicolumn{1}{c|}{0.53}                                                         & \multicolumn{1}{c|}{0.881}                                                           & \multicolumn{1}{c|}{0.764}                                                         & 0.874                                                            \\ \hline
EVENT                                                            & \multicolumn{1}{c|}{386}                                                                                 & 30                                                                                     & \multicolumn{1}{c|}{0.892}       & \multicolumn{1}{c|}{0.15}                                                         & \multicolumn{1}{c|}{0.771}                                                           & \multicolumn{1}{c|}{0.205}                                                        & \multicolumn{1}{c|}{0.779}                                                           & \multicolumn{1}{c|}{0.25}                                                          & 0.842                                                            \\ \hline
PRODUCT                                                          & \multicolumn{1}{c|}{72}                                                                                  & 6                                                                                      & \multicolumn{1}{c|}{0.769}       & \multicolumn{1}{c|}{0}                                                            & \multicolumn{1}{c|}{0.667}                                                           & \multicolumn{1}{c|}{0}                                                            & \multicolumn{1}{c|}{0.363}                                                           & \multicolumn{1}{c|}{0}                                                             & 0.727                                                            \\ \hline
FACILITY                                                         & \multicolumn{1}{c|}{208}                                                                                 & 27                                                                                     & \multicolumn{1}{c|}{0.416}       & \multicolumn{1}{c|}{0}                                                            & \multicolumn{1}{c|}{0.4}                                                             & \multicolumn{1}{c|}{0}                                                            & \multicolumn{1}{c|}{0.272}                                                           & \multicolumn{1}{c|}{0}                                                             & 0.421                                                            \\ \hline
ROLE                                                             & \multicolumn{1}{c|}{485}                                                                                 & 35                                                                                     & \multicolumn{1}{c|}{0.968}       & \multicolumn{1}{c|}{0}                                                            & \multicolumn{1}{c|}{0.361}                                                           & \multicolumn{1}{c|}{0}                                                            & \multicolumn{1}{c|}{0.75}                                                            & \multicolumn{1}{c|}{0}                                                             & 0.925                                                            \\ \hline
NORP                                                             & \multicolumn{1}{c|}{114}                                                                                 & 5                                                                                      & \multicolumn{1}{c|}{1}           & \multicolumn{1}{c|}{0.625}                                                        & \multicolumn{1}{c|}{0.833}                                                           & \multicolumn{1}{c|}{0.416}                                                        & \multicolumn{1}{c|}{0.8}                                                             & \multicolumn{1}{c|}{0.588}                                                         & 1                                                                \\ \hline
TITLE                                                            & \multicolumn{1}{c|}{300}                                                                                 & 4                                                                                      & \multicolumn{1}{c|}{0.8}         & \multicolumn{1}{c|}{0}                                                            & \multicolumn{1}{c|}{0}                                                               & \multicolumn{1}{c|}{0}                                                            & \multicolumn{1}{c|}{0}                                                               & \multicolumn{1}{c|}{0}                                                             & 0.4                                                              \\ \hline
LAW                                                              & \multicolumn{1}{c|}{62}                                                                                  & 5                                                                                      & \multicolumn{1}{c|}{0.2}         & \multicolumn{1}{c|}{0.4}                                                          & \multicolumn{1}{c|}{0.8}                                                             & \multicolumn{1}{c|}{0.5}                                                          & \multicolumn{1}{c|}{0.7}                                                             & \multicolumn{1}{c|}{0.2}                                                           & 0.8                                                              \\ \hline
LANGUAGE                                                         & \multicolumn{1}{c|}{0}                                                                                   & 0                                                                                      & \multicolumn{1}{c|}{0}           & \multicolumn{1}{c|}{0}                                                            & \multicolumn{1}{c|}{0}                                                               & \multicolumn{1}{c|}{0}                                                            & \multicolumn{1}{c|}{0}                                                               & \multicolumn{1}{c|}{0}                                                             & 0                                                                \\ \hline
WORK\_OF\_ART                                                    & \multicolumn{1}{c|}{7}                                                                                   & 2                                                                                      & \multicolumn{1}{c|}{0.5}         & \multicolumn{1}{c|}{0}                                                            & \multicolumn{1}{c|}{0.5}                                                             & \multicolumn{1}{c|}{0}                                                            & \multicolumn{1}{c|}{1}                                                               & \multicolumn{1}{c|}{0}                                                             & 1                                                                \\ \hline
                                                                 & \multicolumn{1}{c|}{}                                                                                    &                                                                                        & \multicolumn{1}{c|}{}            & \multicolumn{1}{c|}{}                                                             & \multicolumn{1}{c|}{}                                                                & \multicolumn{1}{c|}{}                                                             & \multicolumn{1}{c|}{}                                                                & \multicolumn{1}{c|}{}                                                              &                                                                  \\ \hline
Total Entities                                                   & \multicolumn{1}{c|}{6061}                                                                                & 634                                                                                    & \multicolumn{7}{c|}{}                                                                                                                                                                                                                                                                                                                                                                                                                                                                                                                          \\ \hline
\begin{tabular}[c]{@{}c@{}}Overall\\ Micro F1-Score\end{tabular} & \multicolumn{2}{c|}{}                                                                                                                                                                             & \multicolumn{1}{c|}{0.94}        & \multicolumn{1}{c|}{0.211}                                                        & \multicolumn{1}{c|}{0.832}                                                           & \multicolumn{1}{c|}{0.25}                                                         & \multicolumn{1}{c|}{0.844}                                                           & \multicolumn{1}{c|}{0.331}                                                         & 0.883                                                            \\ \hline
\end{tabular}%
}
\caption{Fine-Tuned model performance (based on F1-Score) on spaCy Model, calculated based on validation set for each entity labels. }
\label{tab:entity-label-spacy}
\end{table*}

We investigated the contribution of MEN-Dataset in NER using spaCy \citep{spacy3}.  spaCy  is a free and open source NLP library that supports task like Tokenization, NER, and POS Tagging. spaCy has 84 trained pipeline in several languages and that includes English. A pipeline is a sequence of processing steps applied to text in spaCy. Figure \ref{fig:spacy-pipeline} presents the pipeline steps in spaCy. For English, spaCy has 4 trained pipeline named ``en\_core\_web\_sm'', ``en\_core\_web\_md'', ``en\_core\_web\_lg'', ``en\_core\_web\_tf''. 
\begin{enumerate}
    \item en\_core\_web\_sm: This is a small sized model, that is designed for more memory efficient. The small model does not have any word vector. For NER, the pipeline has been trained with OntoNotes 5.0 \citep{ontonotes-new} and achieved F1-Score of 0.845 \footnote{\href{https://github.com/explosion/spacy-models/commit/3824a1d43a6aedab4f8e6e2e6907b92c62a3c27d}{Model Metadata for en\_core\_web\_sm}}. 
    \item en\_core\_web\_md: This is a medium sized model. The medium model has reduced word vector table with 514k keys, 20k unique vectors. Similar to en\_core\_web\_sm, the NER component is trained with OntoNotes 5.0 \citep{ontonotes-new} and achieved F1-Score of 0.852 \footnote{\href{https://github.com/explosion/spacy-models/commit/3f88431423aac8efc03cb81e1a9eb0da05b1da50}{Model Metadata for en\_core\_web\_md}}. 
    \item en\_core\_web\_lg: This is a medium sized model. The medium model has word vector with 514k keys, 514k unique vectors. For NER, the model has reported to achieve F1-Score of 0.854 \footnote{\href{https://github.com/explosion/spacy-models/commit/63c41a55d2d3223eee92d73e3f502d7d39924f75}{Model Metadata for en\_core\_web\_lg}}. 
    \item en\_core\_web\_tf: This model is based on RoBERTa architecture. For NER, the model has reported to achieve F1-Score of 0.90 \footnote{\href{https://github.com/explosion/spacy-models/commit/434012b940d806e51c8414243bd3dca670231360}{Model Metadata  for en\_core\_web\_trf}}. 
\end{enumerate}

\subsection{Fine-Tuning spaCy NER Model}
\label{ssec: train-spaCy}
spaCy allows us to finetune the model using custom dataset. As described in Section \ref{ssec: spacy-intro}, there are different components in the pipeline.  spaCy supports training the pipeline by disabling or enabling specific components, based on the use case. The MEN-Dataset is split into training (75\%), test (10\%) and validation (15\%) total entities of 5065, 453 and 618 respectively. The splitted dataset is used to fine-tune spaCy NER model. In this experiment, we produced the following spaCy models:
\begin{enumerate}
    \item spacy-blank: We loaded and trained a blank spaCy model, which has not been trained with any dataset before. 
    \item en\_core\_web\_sm, en\_core\_web\_lg, en\_core\_web\_trf: These models have already been pre-trained with English datasets.
    \item en\_core\_web\_sm-FT, en\_core\_web\_lg-FT, en\_core\_web\_trf-FT: The pre-trained model will be fine-tuned with our annotated MEN-Dataset. 
\end{enumerate}
Fine-tuning different spaCy models will allow us to identify the best performing spaCy NER model. To fine-tune, we started by converting our annotated dataset into BIO tagging scheme. Then we generated a Configuration File \footnote{\hyperlink{https://spacy.io/usage/training\#quickstart}{Generate Configuration File}}, which could be used to fine-tune NER model. At the end of training, spaCy will provide us the best trained model (model-best) based on the use case. In our work, since we fine-tuned spaCy for NER, the best model will be selected based on the highest F1-Score achieved on validation set.

\subsection{Result and Analysis}
Table \ref{tab:entity-label-spacy} shows the result of evaluation on the fine-tuned spaCy NER model. To evaluate the performance of NER model, we used our validation set on en\_core\_web\_sm, en\_core\_web\_lg, and en\_core\_web\_trf and consider the results as the baseline. The results were then compared with spacy-blank, en\_core\_web\_sm-FT, en\_core\_web\_lg-FT, and en\_core\_web\_trf-FT. Table \ref{tab:entity-label-spacy} have further details on models performance by different entity labels.

Based on the result, we can observe that spacy-blank has achieved the highest F1-Score (0.94), while en\_core\_web\_sm-FT has achieved the lowest F1-Score (0.832). On average all the fine-tuned spaCy models have shown a significant improvement of +230\%, which outperforms the pre-trained language model. If we compared with fine-tuned pre-trained model, the average F1-Score of spaCy model (F1-Score is 0.88) is higher than the further pre-trained model (F1-Score is 0.81).  These are the findings learnt from the experimental results:
\begin{enumerate}
    \item Across all the fine-tuned spaCy models, there is an significant improvement on F1-Score. On average, the fine-tuned model (when compared with en\_core\_web\_sm and en\_core\_web\_sm-FT) has achieved F1-Score improvement of +401\%. 
    \item All the three base spaCy models (en\_core\_web\_sm, en\_core\_web\_lg, en\_core\_web\_trf) have obtained F1-Score of 0 for entity label PRODUCT, FACILITY, ROLE, TITLE, LANGUAGE, and WORK\_OF\_ART. It is important to take note that the NER component in spaCy model is trained upon OntoNotes 5.0 \citep{ontonotes-new} dataset, and in MEN-Dataset we have followed 10 out of 12 entity labels in OntoNotes. As such, apart from the entities labeled as ROLE and TITLE, the base model was expected to predict entities for the remaining entity labels. However, it failed to do so in our experiment.
    \item Referring back to introduction of Malaysian English (in Section \ref{sec:introduction}), entity label that exhibits morphosyntactic adaptions are TITLE, WORK\_OF\_ART, ORGANIZATION, LOCATION, and FACILITY. For these entity labels, we can see a significant improvement achieved by the fine-tuned NER models.  
\end{enumerate}
As the conclusion, our custom spaCy NER model called spacy-blank has achieved highest F1-Score compared to other fine-tuned spaCy model. Our fine-tuning approach has improved the performance of the spaCy NER model for Malaysian English. This improvement is credited to the development of the MEN-Dataset.

\section{Conclusion}
\label{sec: conclusion}
This paper presents our endeavor to construct \textbf{MEN-Dataset}, a comprehensive Malaysian English news articles, annotated with entities and relations. Our thorough background studies highlight two limitations of the existing state-of-the-art (SOTA) solutions pertaining to Malaysian English. First of all, our literature review shows that there is no quality dataset on Malaysian English available. Secondly, the SOTA Named Entity Recognition tools are not able to fully capture the entities in the Malaysian English news articles. To resolve first limitation, we developed a human-annotated \textbf{MEN-Dataset}. MEN-Dataset contains 6,061 entities and 3,268 relation instances. Meanwhile for second limitation, we fine-tuned the performance of an existing state-of-the-art NER tool, namely spaCy and achieved significant improvement in NER. Fine-tuned on MEN-Dataset, our \textbf{spacy-blank} achieves the highest F1-Score of 0.94. On average, all the fine-tuned spaCy models achieve an improvement of +230\%. For the future work, we will be expanding the dataset using Human-In-The-Loop Annotation while ensuring quality of the dataset is preserved. 

\section{Ethical Consideration}
\label{sec:ethical-consideration}

This paper presents a new dataset, constructed following ethical research practice, as discussed below. 
\begin{enumerate}
    \item \textbf{Intellectual Property and Privacy}. The dataset is constructed using selected news articles published by a few news portals in Malaysia, namely New Straits Times (NST) \footnote{\url{https://www.nst.com.my/}},  Malay Mail \footnote{\url{https://www.malaymail.com/}} and Bernama English \footnote{\url{https://www.bernama.com/en/}}. Our institution's Human Research Ethics Committee was consulted prior to the start of the project.  It was concluded that news media do not require any ethics approval as they were written and published for public consumption.  In fact, analysis of news article is commonly done by commentators.  All of the entity mentions are based on the context of news articles. 
    \item \textbf{Compensation}. The annotators were compensated based on the amount of time they spent on the training, discussions, as well as the number of news articles they annotated. 
\end{enumerate}

\nocite{*}
\section{Bibliographical References}\label{sec:reference}

\bibliographystyle{lrec-coling2024-natbib}
\bibliography{lrec-coling2024-example}

\onecolumn

\appendix

\section{Words with morphosyntactic adaptations}
\label{appendix:morpho-adaptations}
\begin{table}[H]
\begin{tabular}{|l|l|}
\hline
Word                        & Meaning                                                                   \\ \hline
\textit{kenduri}            & refers to feast.                                                          \\ \hline
\textit{iman}               & a PERSON who leads prayers in the mosque.                                 \\ \hline
\textit{ustaz}              & a religious male teacher.                                                 \\ \hline
\textit{bumiputera}         & a Malaysian of indigenous Malay origin.                                   \\ \hline
\textit{orang ali}          & a collective term for the indigenous peoples of Malaysia.                 \\ \hline
\textit{Datuk}              & a title denoting membership of a high  order of chivalry, given to Males. \\ \hline
\textit{Makcik}     & used to call Aunty respectively.                                \\ \hline
\textit{Dewan Negara}       & is the upper house of the Parliament of Malaysia.                         \\ \hline
\textit{Mentari Besar}      & used to refer to important ministers from political party.                \\ \hline
\textit{Yang Di Pertua}     & is a title for the head of state in certain Malay-speaking countries.     \\ \hline
\textit{tiada apa attitude} & refers to does not care attitude.                                         \\ \hline
\textit{Chinese sinseh}     & a honorific term that used to refer "person born before another".         \\ \hline
\textit{pondok school}      & organised religious schools for traditional Islam people.                 \\ \hline
\textit{Tabung Haji Board}  & the Malaysian hajj pilgrims fund board.                                   \\ \hline
\textit{kampung house}      & a typical Malay house in villages.                                        \\ \hline
\textit{Datukship}          & a title denoting membership of a high order of chivalry, given to Males.  \\ \hline
\textit{Johorean}           & A native or inhabitant of Johor in Malaysia.                              \\ \hline
\textit{non halal}          & foods that not to be eaten by those observing Islamic teachings.          \\ \hline
\textit{ang pows}           & a gift of money packed into a red packet.                                 \\ \hline
\textit{bomohs}             & a Malay shaman and traditional medicine practitioner.                     \\ \hline
\textit{cheongsams}         & a dress worn by Chinese peoples/                                          \\ \hline
\end{tabular}
\caption{Refers to the meaning of words, examples with morphosyntactic adaptation.}
\label{tab:example-of-morphosyntactic-adaptation-words}
\end{table}

\newpage
\section{Result of experiment conducted to evaluate performance of NER tools in Malaysian English News}
\label{appendix:experiment-motivation}
\begin{table*}[ht]
\centering
\resizebox{\textwidth}{!}{%
\begin{tabular}{|c|c|cc|cc|cc|cc|}
\hline
API            & \multirow{4}{*}{Total Entity} & \multicolumn{2}{c|}{Malaya}                                                                      & \multicolumn{2}{c|}{Flair}                                                                      & \multicolumn{2}{c|}{Stanza}                                                                     & \multicolumn{2}{c|}{spaCy}                                                                     \\ \cline{1-1} \cline{3-10} 
Model          &                               & \multicolumn{2}{c|}{ontonotes-xlnet}                                                         & \multicolumn{2}{c|}{ner-ontonotes-large}                                                        & \multicolumn{2}{c|}{en-ontonotes-18classes}                                                     & \multicolumn{2}{c|}{en\_core\_web\_trf}                                                            \\ \cline{1-1} \cline{3-10} 
Evaluation     &                               & \multicolumn{1}{c|}{\begin{tabular}[c]{@{}c@{}}Correctly \\ Classified\end{tabular}} & F1-Score & \multicolumn{1}{c|}{\begin{tabular}[c]{@{}c@{}}Correctly \\ Classified\end{tabular}} & F1-Score & \multicolumn{1}{c|}{\begin{tabular}[c]{@{}c@{}}Correctly \\ Classified\end{tabular}} & F1-Score & \multicolumn{1}{c|}{\begin{tabular}[c]{@{}c@{}}Correctly \\ Classified\end{tabular}} & F1-Score \\ \cline{1-1} \cline{3-10} 
Entity Type    &                               & \multicolumn{1}{c|}{}                                                                &          & \multicolumn{1}{c|}{}                                                                &          & \multicolumn{1}{c|}{}                                                                &          & \multicolumn{1}{c|}{}                                                                &          \\ \hline
PER            & 28                            & \multicolumn{1}{c|}{18}                                                              & 0.64     & \multicolumn{1}{c|}{14}                                                              & 0.5      & \multicolumn{1}{c|}{24}                                                              & 0.86     & \multicolumn{1}{c|}{20}                                                              & 0.71     \\ \hline
LOC            & 38                            & \multicolumn{1}{c|}{10}                                                               & 0.26        & \multicolumn{1}{c|}{5}                                                               & 0.13     & \multicolumn{1}{c|}{0}                                                               & 0        & \multicolumn{1}{c|}{15}                                                              & 0.39     \\ \hline
GPE            & 19                            & \multicolumn{1}{c|}{15}                                                              & 0.78        & \multicolumn{1}{c|}{19}                                                              & 1        & \multicolumn{1}{c|}{13}                                                              & 0.68     & \multicolumn{1}{c|}{16}                                                              & 0.84     \\ \hline
EVENT          & 6                             & \multicolumn{1}{c|}{4}                                                               & 0.75     & \multicolumn{1}{c|}{4}                                                               & 0.67     & \multicolumn{1}{c|}{1}                                                               & 0.17     & \multicolumn{1}{c|}{1}                                                               & 0.17     \\ \hline
ORG            & 21                            & \multicolumn{1}{c|}{14}                                                              & 0.67     & \multicolumn{1}{c|}{18}                                                              & 0.86     & \multicolumn{1}{c|}{13}                                                              & 0.62     & \multicolumn{1}{c|}{14}                                                              & 0.67     \\ \hline
FAC            & 10                            & \multicolumn{1}{c|}{6}                                                               & 0.60      & \multicolumn{1}{c|}{5}                                                               & 0.5      & \multicolumn{1}{c|}{3}                                                               & 0.3      & \multicolumn{1}{c|}{6}                                                               & 0.60     \\ \hline
WORK\_OF\_ART  & 1                             & \multicolumn{1}{c|}{0}                                                               & 0        & \multicolumn{1}{c|}{0}                                                               & 0        & \multicolumn{1}{c|}{0}                                                               & 0        & \multicolumn{1}{c|}{1}                                                               & 1        \\ \hline
NORP           & 7                             & \multicolumn{1}{c|}{5}                                                               & 0.71     & \multicolumn{1}{c|}{7}                                                               & 1        & \multicolumn{1}{c|}{2}                                                               & 0.29     & \multicolumn{1}{c|}{3}                                                               & 0.43     \\ \hline
Micro F1-Score & 130                           & \multicolumn{1}{c|}{72}                                                              & 0.55     & \multicolumn{1}{c|}{72}                                                              & 0.55     & \multicolumn{1}{c|}{56}                                                              & 0.43     & \multicolumn{1}{c|}{76}                                                              & 0.58     \\ \hline
\end{tabular}%
}
\caption{Result of experimentation conducted with 30 sentences from Malaysian English News Article using NER Tool}
\label{tab:experiment-motivation}
\end{table*}
*Note: To ensure a better readability of the Table, it has been moved to Appendix.

\newpage
\section{Annotator Details}
\label{appendix:annotator-details}

\begin{table}[hbt!]
\centering
\resizebox{\columnwidth}{!}{%
\begin{tabular}{|c|c|c|c|c|}
\hline
Annotator & \begin{tabular}[c]{@{}c@{}}Educational \\ Background\end{tabular}                                       & \begin{tabular}[c]{@{}c@{}}Proficiency in \\ Bahasa Malaysia\end{tabular} & \begin{tabular}[c]{@{}c@{}}Familiarity of \\ Malaysian English\end{tabular}                                                             & \begin{tabular}[c]{@{}c@{}}Feedback from \\ Assessment\end{tabular}                                                                                           \\ \hline
1         & \begin{tabular}[c]{@{}c@{}}Diploma in \\ Business Administration.\end{tabular}                          & Intermediate                                                              & \begin{tabular}[c]{@{}c@{}}Using Malaysian English as part of \\ day to day communication, and \\ for educational purpose.\end{tabular} & \begin{tabular}[c]{@{}c@{}}Able complete annotate 12/15 \\ news articles provided and correctly \\ annotate them based on guideline \\ provided.\end{tabular} \\ \hline
2         & \begin{tabular}[c]{@{}c@{}}Bachelor of English \\ Language and Linguistics\end{tabular}                 & Intermediate                                                              & \begin{tabular}[c]{@{}c@{}}Using Malaysian English as part of \\ day to day communication, and \\ for educational purpose.\end{tabular} & \begin{tabular}[c]{@{}c@{}}Able complete annotate 13/15 \\ news articles provided and correctly \\ annotate them based on guideline \\ provided.\end{tabular} \\ \hline
3         & \begin{tabular}[c]{@{}c@{}}Bachelor of Human Science \\ (English Language and Literature).\end{tabular} & Intermediate                                                              & \begin{tabular}[c]{@{}c@{}}Using Malaysian English as part of \\ day to day communication, and \\ for educational purpose.\end{tabular} & \begin{tabular}[c]{@{}c@{}}Able complete annotate 15/15 \\ news articles provided and correctly \\ annotate them based on guideline \\ provided.\end{tabular} \\ \hline
4         & Bachelor of English Literature                                                                          & Intermediate                                                              & \begin{tabular}[c]{@{}c@{}}Using Malaysian English as part of \\ day to day communication, and \\ for educational purpose.\end{tabular} & \begin{tabular}[c]{@{}c@{}}Able complete annotate 15/15 \\ news articles provided and correctly \\ annotate them based on guideline \\ provided.\end{tabular} \\ \hline
\end{tabular}%
}
\caption{Details of annotators selected for annotation work.}
\label{tab:annotator-details}
\end{table}

\newpage
\section{List of Named Entity labels}
\label{appendix:list-entity-labels}
\begin{table}[hbt!]
\centering
\begin{tabular}{|l|l|l|}
\hline
No & Entity Label & Description                                                                                                                                                                                                                                    \\ \hline
1  & PERSON       & \begin{tabular}[c]{@{}l@{}}The Entity PERSON includes Name of Person in the text. \\ This entity type has been adapted from OntoNotes 5.0.\end{tabular}                                                                                        \\ \hline
2  & LOCATION     & \begin{tabular}[c]{@{}l@{}}LOCATION is any place that can be occupied by or has been \\ occupied by someone in this EARTH and outside of EARTH. \\ Entity mention that could be labelled as GPE has been labelled \\ as LOCATION.\end{tabular} \\ \hline
3  & ORGANIZATION & ORGANIZATION is group of people with specific purpose.                                                                                                                                                                                         \\ \hline
4  & NORP         & \begin{tabular}[c]{@{}l@{}}NORP is the abbrevation for the term Nationality, Religious \\ or Political group.\end{tabular}                                                                                                                     \\ \hline
5  & FACILITY     & FACILITY refers to man-made structures.                                                                                                                                                                                                        \\ \hline
6  & PRODUCT      & \begin{tabular}[c]{@{}l@{}}PRODUCT refers to an object, or a service that is made \\ available for consumer use as of the consumer demand.\end{tabular}                                                                                        \\ \hline
7  & EVENT        & \begin{tabular}[c]{@{}l@{}}An EVENT is a reference to an organized or unorganized \\ incident.\end{tabular}                                                                                                                                    \\ \hline
8  & WORK OF ART  & \begin{tabular}[c]{@{}l@{}}WORK OF ART refers to ART entities that has been made \\ by a PERSON or ORGANIZATION.\end{tabular}                                                                                                                  \\ \hline
9  & LAW          & \begin{tabular}[c]{@{}l@{}}LAW are rules that has been made by an authority and that\\ must be obeyed.\end{tabular}                                                                                                                            \\ \hline
10 & LANGUAGE     & LANGUAGE refers to any named language.                                                                                                                                                                                                         \\ \hline
11 & ROLE         & \begin{tabular}[c]{@{}l@{}}ROLE is used to define the position or function of the \\ PERSON in an ORGANIZATION.\end{tabular}                                                                                                                   \\ \hline
12 & TITLE        & TITLE is used to define the honorific title of the PERSON.                                                                                                                                                                                     \\ \hline
\end{tabular}
\caption{List of 12 entity labels and description.}
\label{tab:entity-types}
\centering
\end{table}

\newpage
\begin{landscape}
\section{List of Relation labels}
\label{appendix:list-relation-labels}
\begin{longtable}{| p{.025\textwidth} | p{.20\textwidth} | p{.10\textwidth} | p{.20\textwidth} | p{.20\textwidth} | p{.60\textwidth} |} 
\hline
No & Relation Label & Dataset Adapted & Entity Type One & Entity Type Two & Description \\ \hline
1 & head of government & DocRED & PER & ORG,LOC & head of the executive power of this town, city, municipality, state, country, or other governmental body \\ \hline
2 & country & DocRED & PER,ORG & LOC & sovereign state of this item (not to be used for human beings) \\ \hline
3 & place of birth & DocRED & PER & LOC & most specific known (e.g. city instead of country, or hospital instead of city) birth location of a person, animal or fictional character \\ \hline
4 & place of death & DocRED & PER & LOC & most specific known (e.g. city instead of country, or hospital instead of city) death location of a person, animal or fictional character \\ \hline
5 & father & DocRED & PER & PER & "male parent of the subject." \\ \hline
6 & mother & DocRED & PER & PER & "female parent of the subject." \\ \hline
7 & spouse & DocRED & PER & PER & "the subject has the object as their spouse (husband, wife, partner, etc.)." \\ \hline
8 & country of citizenship & DocRED & LOC & PER & the object is a country that recognizes the subject as its citizen \\ \hline
9 & continent & DocRED & LOC & LOC & continent of which the subject is a part \\ \hline
10 & head of state & DocRED & PER & LOC & official with the highest formal authority in a country/state \\ \hline
11 & capital & DocRED & LOC & LOC & seat of government of a country, province, state or other type of administrative territorial entity \\ \hline
12 & official language & DocRED & LOC,ORG & PER & language designated as official by this item \\ \hline
13 & position held & DocRED & PER & ROLE & subject currently or formerly holds the object position or public office \\ \hline
14 & child & DocRED & PER & PER & subject has object as child. Do not use for stepchildren \\ \hline
15 & author & DocRED & PER & WORK\_OF\_ART & main creator(s) of a written work \\ \hline
16 & director & DocRED & PER & WORK\_OF\_ART & director(s) of film, TV-series, stageplay, video game or similar \\ \hline
17 & screenwriter & DocRED & PER & WORK\_OF\_ART & person(s) who wrote the script for subject item \\ \hline
18 & educated at & DocRED & PER & ORG & educational institution attended by subject \\ \hline
19 & composer & DocRED & PER & WORK\_OF\_ART & "person(s) who wrote the music" \\ \hline
20 & occupation & DocRED & PER & ROLE & "occupation of a person" \\ \hline
21 & founded by & DocRED & PER & ORG & founder or co-founder of this organization, religion or place \\ \hline
22 & league & DocRED & ORG & EVENT & league in which team or player plays or has played in \\ \hline
23 & place of burial & DocRED & PER & LOC & location of grave, resting place, place of ash-scattering, etc. (e.g., town/city or cemetery) for a person or animal. There may be several places: e.g., re-burials, parts of body buried separately. \\ \hline
24 & publisher & DocRED & PER & WORK\_OF\_ART & organization or person responsible for publishing books, periodicals, printed music, podcasts, games or software \\ \hline
25 & maintained by & DocRED & PER,ORG & FAC,ORG & person or organization in charge of keeping the subject (for instance an infrastructure) in functioning order \\ \hline
26 & owned by & DocRED & PER & ORG, FAC, PRODUCT & owner of the subject \\ \hline
27 & operator & DocRED & PER & PRODUCT,FAC & person, profession, or organization that operates the equipment, facility, or service \\ \hline
28 & named after & DocRED & PER & FAC,ORG,EVENT & "entity or event that inspired the subject's name, or namesake (in at least one language)." \\ \hline
29 & cast member & DocRED & PER & WORK\_OF\_ART & "actor in the subject production" \\ \hline
30 & producer & DocRED & PER & WORK\_OF\_ART & person(s) who produced the film, musical work, theatrical production, etc. (for film, this does not include executive producers, associate producers, etc.) \\ \hline
31 & award received & DocRED & PER, ORG, WORK\_OF\_ART, TITLE & WORK\_OF\_ART, TITLE & award or recognition received by a person, organization or creative work \\ \hline
32 & chief executive officer & DocRED & PER & ORG & highest-ranking corporate officer appointed as the CEO within an organization \\ \hline
33 & creator & DocRED & PER & WORK\_OF\_ART, PRODUCT & maker of this creative work or other object (where no more specific property exists) \\ \hline
34 & ethnic group & DocRED & PER & ORG & subject's ethnicity (consensus is that a VERY high standard of proof is needed for this field to be used. In general this means 1) the subject claims it themselves, or 2) it is widely agreed on by scholars, or 3) is fictional and portrayed as such) \\ \hline
35 & performer & DocRED & PER & WORK\_OF\_ART & actor, musician, band or other performer associated with this role or musical work \\ \hline
36 & manufacturer & DocRED & ORG & PRODUCT & manufacturer or producer of this product \\ \hline
37 & developer & DocRED & ORG,PER & PRODUCT,FAC & organization or person that developed the item \\ \hline
38 & legislative body & DocRED & ORG & ORG & legislative body governing this entity; political institution with elected representatives, such as a parliament/legislature or council \\ \hline
39 & executive body & DocRED & ORG & ORG & branch of government for the daily administration of the territorial entity \\ \hline
40 & record label & DocRED & ORG & WORK\_OF\_ART & brand and trademark associated with the marketing of subject music recordings and music videos \\ \hline
41 & production company & DocRED & ORG & WORK\_OF\_ART & company that produced this film, audio or performing arts work \\ \hline 
42 & location & DocRED & PER,FAC,ORG & LOC & location of the object, structure or event. \\ \hline
43 & place of publication & DocRED & WORK\_OF\_ART & LOC & geographical place of publication of the edition (use 1st edition when referring to works) \\ \hline
44 & part of & DocRED & PER & ORG,EVENT & "object of which the subject is a part (if this subject is already part of object A which is a part of object B, then please only make the subject part of object A)." \\ \hline
45 & military rank & DocRED & PER & ROLE & "military rank achieved by a person (should usually have a ""start time"" qualifier), or military rank associated with a position" \\ \hline
46 & member of & DocRED & PER & ORG & organization, club or musical group to which the subject belongs. Do not use for membership in ethnic or social groups, nor for holding a political position, such as a member of parliament. \\ \hline
47 & chairperson & DocRED & PER & ORG & presiding member of an organization, group or body \\ \hline
48 & country of origin & DocRED & LOC & WORK\_OF\_ART, PRODUCT & country of origin of this item (creative work, food, phrase, product, etc.) \\ \hline
49 & diplomatic relation & DocRED & ORG & ORG & diplomatic relations of the country \\ \hline
50 & residence & DocRED & PER & FAC,LOC & the place where the person is or has been, resident \\ \hline
51 & organizer & DocRED & PER,ORG & EVENT & person or institution organizing an event \\ \hline
52 & characters & DocRED & PER & WORK\_OF\_ART & characters which appear in this item (like plays, operas, operettas, books, comics, films, TV series, video games) \\ \hline
53 & lyrics by & DocRED & PER & WORK\_OF\_ART & author of song lyrics \\ \hline
54 & participant & DocRED & PER,ORG & EVENT,ORG & "person, group of people or organization (object) that actively takes/took part in an event or process (subject)." \\ \hline
55 & given name & DocRED & PER & PER & first name or another given name of this person; values used with the property should not link disambiguations nor family names \\ \hline
56 & location of formation & DocRED & ORG & LOC & location where a group or organization was formed \\ \hline
57 & parent organization & DocRED & ORG & ORG & parent organization of an organization. \\ \hline
58 & significant event & DocRED & PER,ORG & EVENT & significant or notable events associated with the subject \\ \hline
59 & authority & DocRED & PER & ORG & entity having executive power on given entity \\ \hline
60 & sponsor & DocRED & PER,ORG & PER,EVENT & organization or individual that sponsors this item \\ \hline
61 & applies to jurisdiction & DocRED & LAW & LOC & the item (institution, law, public office, public register...) or statement belongs to or has power over or applies to the value (a territorial jurisdiction: a country, state, municipality, ...) \\ \hline
62 & director / manager & DocRED & PER & ORG & person who manages any kind of group \\ \hline
63 & product or material produced & DocRED & PER & WORK\_OF\_ART & material or product produced by a government agency, business, industry, facility, or process \\ \hline
64 & student of & DocRED & PER & PER & person who has taught this person \\ \hline
65 & territory claimed by & DocRED & ORG & LOC & administrative divisions that claim control of a given area \\ \hline
66 & winner & DocRED & PER,ORG & EVENT & "winner of a competition or similar event, not to be used for awards or for wars or battles" \\ \hline
67 & replaced by & DocRED & PER & PER & "other person or item which continues the item by replacing it in its role." \\ \hline
68 & capital of & DocRED & LOC & LOC & country, state, department, canton or other administrative division of which the municipality is the governmental seat \\ \hline
69 & languages spoken, written or signed & DocRED & PER & LANGUAGE & language(s) that a person or a people speaks, writes or signs, including the native language(s) \\ \hline
70 & present in work & DocRED & PER & WORK\_OF\_ART & this (fictional or fictionalized) entity or person appears in that work as part of the narration \\ \hline
71 & country for sport & DocRED & PER,ORG & LOC & country a person or a team represents when playing a sport \\ \hline
72 & represented by & DocRED & PER & ORG & person or agency that represents or manages the subject \\ \hline
73 & investor & DocRED & PER,ORG & ORG & individual or organization which invests money in the item for the purpose of obtaining financial return on their investment \\ \hline
74 & intended public & DocRED & PER,ORG & PRODUCT,EVENT & this work, product, object or event is intended for, or has been designed to that person or group of people, animals, plants, etc \\ \hline
75 & partnership with & DocRED & ORG & ORG & partnership (commercial or/and non-commercial) between this organization and another organization or institution \\ \hline
76 & statistical leader & DocRED & ORG,PER & EVENT & leader of a sports tournament in one of statistical qualities (points, assists, rebounds etc.). \\ \hline
77 & board member & DocRED & PER & ORG & member(s) of the board for the organization \\ \hline
78 & sibling & DocRED & PER & PER & "the subject and the object have at least one common parent (brother, sister, etc. including half-siblings)" \\ \hline
79 & stepparent & DocRED & PER & PER & subject has the object as their stepparent \\ \hline
80 & candidacy in election & DocRED & PER,ORG & EVENT & election where the subject is a candidate \\ \hline
81 & coach of sports team & DocRED & PER & ORG & sports club or team for which this person is or was on-field manager or coach \\ \hline
82 & subsidiary & DocRED & ORG & ORG & subsidiary of a company or organization; generally a fully owned separate corporation. \\ \hline
83 & religion & DocRED & PER & ORG & religion of a person, organization or religious building, or associated with this subject \\ \hline
84 & Physical.Located & ACE-2005 & PER & FAC, LOC & Located captures the physical location of an entity. \\ \hline
85 & Physical.Near & ACE-2005 & PER, FAC, LOC & FAC, LOC & Indicates that an entity is explicitly near another entity. \\ \hline
86 & Part-Whole.Geo & ACE-2005 & FAC, LOC & FAC, LOC & Captures the location of FAC, LOC, or GPE in or at or as a part of another FAC, LOC or GPE. \\ \hline
87 & Part-Whole.Subsidary & ACE-2005 & ORG & ORG, LOC & Captures the ownership, administrative, and other hierarchical relationships between organizations and between organizations and GPEs. \\ \hline
88 & Per-Social.Business & ACE-2005 & PER & PER & Captures the connection between two entities in any professional relationships. \\ \hline
89 & Per-Social.Family & ACE-2005 & PER & PER & Captures the connection between one entity and another entity in family relations. \\ \hline
90 & Per-Social.Lasting & ACE-2005 & PER & PER & Captures the relations that invlovle personal contact (Where one entity has spent time with another entity, like classmate, neighbor), or indication that the relationships exists outside of a particular cited interaction. \\ \hline
91 & Org-Aff.Employment & ACE-2005 & PER & ORG,LOC & Captures relationship between Person and their employers. \\ \hline
92 & Org-Aff.Ownership & ACE-2005 & PER & ORG & Captures relationship between a Person and an Organization owned by that PERSON \\ \hline
93 & Org-Aff.Founder & ACE-2005 & PER,ORG & ORG,LOC & Captures relation between an entity and an organization that has been founded by the entity \\ \hline
94 & Org-Aff.Student-Alum & ACE-2005 & PER & ORG-Educational ONLY & Captures relation between Person and an educational institution. \\ \hline
95 & Org-Aff.Sports-Affiliation & ACE-2005 & PER & ORG & Captures relation between Player, Coach, Manager with their affiliated Sport ORG \\ \hline
96 & Org-Aff.Shareholder & ACE-2005 & PER, ORG, GPE & ORG, GPE & Captures the relation between an agent and an Organization \\ \hline
97 & Org-Aff.Membership & ACE-2005 & PER, ORG, GPE & ORG & Membership captures relation between an entity and organization which the entity is a member of \\ \hline
98 & Agent-Artifact.UOIM & ACE-2005 & PER, ORG, GPE & FAC & When an entity own an artifact, uses an artifact or caused an artifact to come into being. \\ \hline
99 & Gen-Aff.CRRE & ACE-2005 & PER & ORG, LOC & "When there is a relation between PER and LOC in which they have citizenship. Or when there is a relation between PER and LOC they live. Or when when there is a relation between PER and religious ORG or PER. Or when there is a relation between PER and LOC or PER entity that indicates their ethnicity"  \\ \hline
100 & Gen-Aff.Loc-Origin & ACE-2005 & ORG & LOC & Captures the relation between an organization and the LOC where it is located. \\ \hline
101 & NO\_RELATION & ~ & ANY ENTITY & ANY ENTITY & Can be used for any entity pair that does not have a suitable Relations Listed \\ \hline
\caption{List of 101 relation labels and description.}
\label{tab:relation-labels}\\
\end{longtable}
\end{landscape}

\newpage
\begin{landscape}
\section{Performance of Fine-tuned spaCy Model based on Entity Label}
\label{appendix:entity-label-spacy}
\begin{longtable}[c]{|c|cc|cccccll|}
\hline
Entity Label                                                     & \multicolumn{1}{c|}{\begin{tabular}[c]{@{}c@{}}Total Annotated \\ Entity in \\ MEN-Dataset\end{tabular}} & \begin{tabular}[c]{@{}c@{}}Total Annotated \\ Entity in \\ Validation Set\end{tabular} & \multicolumn{1}{c|}{spacy-blank} & \multicolumn{1}{c|}{\begin{tabular}[c]{@{}c@{}}en\_core\_\\ web\_sm\end{tabular}} & \multicolumn{1}{c|}{\begin{tabular}[c]{@{}c@{}}en\_core\_\\ web\_sm-FT\end{tabular}} & \multicolumn{1}{c|}{\begin{tabular}[c]{@{}c@{}}en\_core\_\\ web\_lg\end{tabular}} & \multicolumn{1}{c|}{\begin{tabular}[c]{@{}c@{}}en\_core\_\\ web\_lg-FT\end{tabular}} & \multicolumn{1}{c|}{\begin{tabular}[c]{@{}c@{}}en\_core\_\\ web\_trf\end{tabular}} & \begin{tabular}[c]{@{}c@{}}en\_core\_\\ web\_trf-FT\end{tabular} \\ \hline
\endfirsthead
\endhead
PERSON                                                           & \multicolumn{1}{c|}{1646}                                                                                & 108                                                                                    & \multicolumn{1}{c|}{0.982}       & \multicolumn{1}{c|}{0.39}     & \multicolumn{1}{c|}{0.852}        & \multicolumn{1}{c|}{0.555}    & \multicolumn{1}{c|}{0.806}        & \multicolumn{1}{l|}{0.811}     & 0.923                      \\ \hline
LOCATION                                                         & \multicolumn{1}{c|}{1157}                                                                                & 150                                                                                    & \multicolumn{1}{c|}{0.942}       & \multicolumn{1}{c|}{0.04}     & \multicolumn{1}{c|}{0.897}        & \multicolumn{1}{c|}{0.041}    & \multicolumn{1}{c|}{0.906}        & \multicolumn{1}{l|}{0.043}     & 0.916                      \\ \hline
ORGANIZATION                                                     & \multicolumn{1}{c|}{1624}                                                                                & 262                                                                                    & \multicolumn{1}{c|}{0.956}       & \multicolumn{1}{c|}{0.474}    & \multicolumn{1}{c|}{0.884}        & \multicolumn{1}{c|}{0.53}     & \multicolumn{1}{c|}{0.881}        & \multicolumn{1}{l|}{0.764}     & 0.874                      \\ \hline
EVENT                                                            & \multicolumn{1}{c|}{386}                                                                                 & 30                                                                                     & \multicolumn{1}{c|}{0.892}       & \multicolumn{1}{c|}{0.15}     & \multicolumn{1}{c|}{0.771}        & \multicolumn{1}{c|}{0.205}    & \multicolumn{1}{c|}{0.779}        & \multicolumn{1}{l|}{0.25}      & 0.842                      \\ \hline
PRODUCT                                                          & \multicolumn{1}{c|}{72}                                                                                  & 6                                                                                      & \multicolumn{1}{c|}{0.769}       & \multicolumn{1}{c|}{0}        & \multicolumn{1}{c|}{0.667}        & \multicolumn{1}{c|}{0}        & \multicolumn{1}{c|}{0.363}        & \multicolumn{1}{l|}{0}         & 0.727                      \\ \hline
FACILITY                                                         & \multicolumn{1}{c|}{208}                                                                                 & 27                                                                                     & \multicolumn{1}{c|}{0.416}       & \multicolumn{1}{c|}{0}        & \multicolumn{1}{c|}{0.4}          & \multicolumn{1}{c|}{0}        & \multicolumn{1}{c|}{0.272}        & \multicolumn{1}{l|}{0}         & 0.421                      \\ \hline
ROLE                                                             & \multicolumn{1}{c|}{485}                                                                                 & 35                                                                                     & \multicolumn{1}{c|}{0.968}       & \multicolumn{1}{c|}{0}        & \multicolumn{1}{c|}{0.361}        & \multicolumn{1}{c|}{0}        & \multicolumn{1}{c|}{0.75}         & \multicolumn{1}{l|}{0}         & 0.925                      \\ \hline
NORP                                                             & \multicolumn{1}{c|}{114}                                                                                 & 5                                                                                      & \multicolumn{1}{c|}{1}           & \multicolumn{1}{c|}{0.625}    & \multicolumn{1}{c|}{0.833}        & \multicolumn{1}{c|}{0.416}    & \multicolumn{1}{c|}{0.8}          & \multicolumn{1}{l|}{0.588}     & 1                          \\ \hline
TITLE                                                            & \multicolumn{1}{c|}{300}                                                                                 & 4                                                                                      & \multicolumn{1}{c|}{0.8}         & \multicolumn{1}{c|}{0}        & \multicolumn{1}{c|}{0}            & \multicolumn{1}{c|}{0}        & \multicolumn{1}{c|}{0}            & \multicolumn{1}{l|}{0}         & 0.4                        \\ \hline
LAW                                                              & \multicolumn{1}{c|}{62}                                                                                  & 5                                                                                      & \multicolumn{1}{c|}{0.2}         & \multicolumn{1}{c|}{0.4}      & \multicolumn{1}{c|}{0.8}          & \multicolumn{1}{c|}{0.5}      & \multicolumn{1}{c|}{0.7}          & \multicolumn{1}{l|}{0.2}       & 0.8                        \\ \hline
LANGUAGE                                                         & \multicolumn{1}{c|}{0}                                                                                   & 0                                                                                      & \multicolumn{1}{c|}{0}           & \multicolumn{1}{c|}{0}        & \multicolumn{1}{c|}{0}            & \multicolumn{1}{c|}{0}        & \multicolumn{1}{c|}{0}            & \multicolumn{1}{l|}{0}         & 0                          \\ \hline
WORK\_OF\_ART                                                    & \multicolumn{1}{c|}{7}                                                                                   & 2                                                                                      & \multicolumn{1}{c|}{0.5}         & \multicolumn{1}{c|}{0}        & \multicolumn{1}{c|}{0.5}          & \multicolumn{1}{c|}{0}        & \multicolumn{1}{c|}{1}            & \multicolumn{1}{l|}{0}         & 1                          \\ \hline
                                                                 & \multicolumn{1}{c|}{}                                                                                    &                                                                                        & \multicolumn{1}{c|}{}            & \multicolumn{1}{c|}{}         & \multicolumn{1}{c|}{}             & \multicolumn{1}{c|}{}         & \multicolumn{1}{c|}{}             & \multicolumn{1}{l|}{}          &                            \\ \hline
Total Entities                                                   & \multicolumn{1}{c|}{6061}                                                                                & 634                                                                                    & \multicolumn{7}{l|}{}                                                                                                                                                                                                                  \\ \hline
\begin{tabular}[c]{@{}c@{}}Overall\\ Micro F1-Score\end{tabular} & \multicolumn{2}{c|}{}                                                                                                                                                                             & \multicolumn{1}{r|}{0.94}        & \multicolumn{1}{r|}{0.211}    & \multicolumn{1}{r|}{0.832}        & \multicolumn{1}{r|}{0.25}     & \multicolumn{1}{r|}{0.844}        & \multicolumn{1}{r|}{0.331}     & \multicolumn{1}{r|}{0.883} \\ \hline
\caption{Fine-Tuned model performance (based on F1-Score) on spaCy Model, calculated based on validation set for each entity labels. }
\label{tab:entity-label-spacy}\\
\end{longtable}
\end{landscape}
\end{document}